\documentclass{article}

\usepackage{arxiv}

\usepackage[utf8]{inputenc} % allow utf-8 input
\usepackage[T1]{fontenc}    % use 8-bit T1 fonts
\usepackage{hyperref}       % hyperlinks
\usepackage{url}            % simple URL typesetting
\usepackage{booktabs}       % professional-quality tables
\usepackage{amsfonts}       % blackboard math symbols
\usepackage{nicefrac}       % compact symbols for 1/2, etc.
\usepackage{microtype}      % microtypography
\usepackage{lipsum}
\usepackage{graphicx}
\graphicspath{ {./images/} }

\usepackage{cite}
\usepackage{amsmath,amssymb,amsfonts}
\usepackage{graphicx}
\usepackage{textcomp}
\usepackage{xcolor}
\usepackage{color}
\newcommand{\ignore}[1]{}
\usepackage{algorithm}
\usepackage[noend]{algorithmic}

\usepackage{multirow}

\author{
Thilina Pathirage Don\\
Optimisation and Logistics\\
School of Computer and Mathematical Sciences\\
The University of Adelaide\\
Adelaide, Australia
\And
Aneta Neumann\\
Optimisation and Logistics\\
School of Computer and Mathematical Sciences\\
The University of Adelaide\\
Adelaide, Australia
\And
Frank Neumann\\
Optimisation and Logistics\\
School of Computer and Mathematical Sciences\\
The University of Adelaide\\
Adelaide, Australia
}

\title{Weighted-Scenario Optimisation for the Chance Constrained Travelling Thief Problem}

\begin{document}
\maketitle

\begin{abstract}
The chance constrained travelling thief problem (chance constrained TTP) has been introduced as a stochastic variation of the classical travelling thief problem (TTP) in an attempt to embody the effect of uncertainty in the problem definition. In this work, we characterise the chance constrained TTP using a limited number of weighted scenarios. Each scenario represents a similar TTP instance, differing slightly in the weight profile of the items and associated with a certain probability of occurrence. Collectively, the weighted scenarios represent a relaxed form of a stochastic TTP instance where the objective is to maximise the expected benefit while satisfying the knapsack constraint with a larger probability. We incorporate a set of evolutionary algorithms and heuristic procedures developed for the classical TTP, and formulate adaptations that apply to the weighted scenario-based representation of the problem. The analysis focuses on the performance of the algorithms on different settings and examines the impact of uncertainty on the quality of the solutions.
\end{abstract}
\keywords{Chance constraints \and scenario-based optimisation \and weighted-sampling \and travelling thief problem.}

\section{Introduction}
The uncertainty of the problem components is a major factor in modelling and optimising real-world problems. 
Relying on deterministic models can limit the alignment between theoretical solutions and real-world conditions.
Therefore, incorporating inter-dependencies among the problem components and integrating stochastic components into the mathematical representations can lead to more realistic optimisation of the real-world scenarios.

The travelling thief problem (TTP)~\cite{bonyadi2013travelling} is a well-regarded multi-component optimisation problem that integrates two conflicting and interdependent components into a single objective function.
The introduction of TTP has drawn attention to modelling more realistic representations of real-world issues, as the essence of certain real-world applications tends to consist of multiple intertwined components.
The chance constrained travelling thief problem (chance constrained TTP)~\cite{PathirageDon2024}, which contains stochastic weights, encapsulates both the uncertainty and interdependency aspects in the problem formulation.
The chance constrained TTP look for the combination of a tour and a packing plan that maximises the expected benefit, meeting the knapsack constraint with a larger probability.
In the literature, the chance constrained TTP has been discussed, representing the problem using surrogate-based and sampling-based models~\cite{PathirageDon2024}.
The presence of stochastic components in an optimisation problem significantly increases the complexity of reaching optimal solutions. However, when finding solid answers to a real-world problem, the expectation is to acquire objectives regardless of the uncertainty imposed by the environmental conditions. Securing the robustness required to withstand every possible scenario is extremely challenging. As a relaxed strategy, the optimisation could be based on sampling a finite number of scenarios that represent the stochastic nature of the problem.

In this work, we focus on optimising the chance constrained TTP by representing the stochastic nature of the problem using a set of weighted scenarios.
Each scenario contains a different weight profile, where the weights are obtained from a given distribution, and the scenario is associated with a probability of occurrence.
The objective would be to maximise the expected fitness value, subject to the knapsack constraint that needs to be satisfied with a given probability. 
Solving the scenario-based representation of the TTP produces a set of high-quality individuals, which collectively provides a relaxed solution to tackle the stochastic problem.

\subsection{Related Work}
Since the introduction of the TTP~\cite{bonyadi2013travelling}, a diverse range of research has been carried out, and a few different variations of the problem have also been investigated. The key contributions include the initial sequential approaches to solve the two components~\cite{polyakovskiy2014comprehensive, mei2014improving}, co-evolutionary algorithms \cite{bonyadi2014socially}, dynamic programming and constraint programming~\cite{wu2017exact}, simulated annealing~\cite{yafrani2018efficiently}, genetic algorithms~\cite{wuijts2019investigation}, ant colony optimisation~\cite{zouari2019new}, restart-based optimisation~\cite{namazi2020surrogate}, quality diversity~\cite{Nikfarjam2024}, coordination based approaches~\cite{namazi2023solving}, local optima networks~\cite{yafrani2022fitness} and a variety of heuristic approaches \cite{faulkner2015approximate, yafrani2016population, namazi2019profit, maity2020efficient}. Besides that, few studies addressed the TTP on stochastic~\cite{PathirageDon2024} and dynamic grounds~\cite{sachdeva2020dynamic}.

Chance constrained optimisation~\cite{miller1965chance} has been studied related to different optimisation problems including submodular functions~\cite{doerr2020optimization,yan2023optimizing,Neumann2020,neumann2024optimizing, DBLP:conf/gecco/NeumannB021,POCCMSP}, the knapsack problem~\cite{xie2019evolutionary,xie2020specific, assimi2020evolutionary,DBLP:conf/ppsn/NeumannXN22, neumann2019runtime, DBLP:conf/gecco/XieN0S21,DBLP:conf/gecco/XieN0S21,DBLP:conf/ppsn/NeumannXN22,MOEAwSWS,E23OMOEA,U3OEADCHKP}, the vehicle routing problem~\cite{kohout1999time, bent2003dynamic, bent2007waiting}, the makespan scheduling problem~\cite{shi2022runtime}, the multiple-choice knapsack problem~\cite{Li2024} and the travelling thief problem~\cite{PathirageDon2024}.
In the literature, sampling-based methods and surrogate-based techniques have been mainly experimented in evaluating chance-constrained optimisation problems~\cite{xie2019evolutionary, PathirageDon2024}.

\subsection{Our Contribution}

In this study, we introduce a chance constrained variation of the TTP relating to a weighted scenario-based model. We define the problem as relating to a limited number of weighted scenarios, which share the same TSP component, yet a different KP component.
After that, we refer to the standard TTP instances and modify them to generate instances that represent $k$ scenarios to fit into the weighted scenario-based model. In order to test the model, we refer to a few well-known evolutionary and heuristic algorithms and introduce adopted versions to evaluate this specific problem. Finally, we deliver a comprehensive analysis of the algorithmic behaviour in this uncertain environment captured via the weighted scenario-based model.

The rest of the sections are structured as follows. Section~\ref{2} provides background on the TTP with a note on applying chance constraints for the TTP. Section~\ref{3} defines the weighted scenario-based model and provides mathematical formulations for the objective function and the chance constraint. After that, Section~\ref{4} describes the adopted algorithms under two subsections, as evolutionary algorithms and local search heuristics. Section~\ref{5} explains the experimental setup and discusses the results that we obtain. Finally, Section~\ref{6} provides the concluding remarks.

\section{Background}
\label{2}

\subsection{Travelling Thief Problem}
\label{2_1}
The TTP involves a set of items, with a profit and a weight, distributed across a given number of cities, except for the first city. A thief (or an agent) starting from the first city, visits each city exactly once, collects items into a knapsack with a predefined capacity, and returns to the starting point. A TTP solution consists of two components: a tour and a packing plan. The tour defines the thief’s route, ensuring each city is visited once before returning. The packing plan specifies the items collected along the way. An objective function calculates the objective score by retrieving the difference between the total profit earned through the collected items and the total travel cost. The travel cost is based on the total weight carried through the tour. The optimisation goal of the TTP is to find out the tour and the packing plan that maximises the objective score without exceeding the knapsack capacity.

The classical version of the TTP has been introduced in~\cite{bonyadi2013travelling} and can be stated as follows. Given a set of items $M=\{e_1, \ldots, e_m\}$, where each item $e_i$ has a profit $p_i$ and a weight $w_i$, $1 \leq i \leq m$, and a set of cities $N=\{1, \ldots, n\}$ with distances $d_{ij}$, $i, j \in N$, between them, the goal is to determine an optimal combination of a permutation of the cities $x$ together with a selection of items $y$.

Located at city $i$ is a set of items $M_i$, $2 \leq i \leq n$, and we have $M= \cup_{i \in N} M_i$. The thief starts the tour from city $1$, visits every city once, collects items while travelling and returns to city $1$. The speed of the thief depends on the weight of the collected items and affects the total travel time. 

Formally, the goal is to find a solution $r=(x,y)$ that consists of a tour (permutation of the cities) $x = ( x_1, \ldots, x_n ), x_i \in N$ and a packing plan $y = ( y_1, \ldots, y_m) \in \{ 0,1 \}^m$ which maximises
\begin{equation}
\label{alg: classical_ttp}
\small
z(x,y) = g(y) - R \left( \frac{d_{x_n x_1}}{\nu_{max}-\nu W_{x_n}} + \sum_{i=1}^{n-1} \frac{d_{x_i x_{i+1}}}{\nu_{max}-\nu W_{x_i}} \right)   
\end{equation}
\begin{equation}
\label{classical TTP - constraint}
\text{subject to }  \sum_{j=1}^{m} w_j y_j \leq B.
\end{equation}

Here $g(y)=\sum_{j=1}^{m} p_{j} y_{j}$ is the total profit of the collected items (the KP solution) and Equation~\ref{classical TTP - constraint} is a classical knapsack constraint which means that the total weight of the collected items can not exceed a given weight bound $B$. The parameters $\nu_{\max}$ and $\nu_{\min}$ are the maximum and minimum traveling speed and $\nu = \frac{\nu_{max}-\nu_{min}}{B}$ is a normalizing constant. $W_{x_i}$ is the accumulated weight of the items collected from city $x_1$ to city $x_i$ in the given tour $x$. A cost factor $R$ called the renting rate is applied to the overall travel time,

\subsection{The Chance Constrained Travelling Thief Problem}
\label{2_2}
The chance constrained TTP is similar to the deterministic TTP, except for the weights of the items, which are not deterministic but stochastic. Our problem formulation incorporates this uncertainty through a probabilistic constraint, which is also known as a chance constraint. Hence, to obtain the feasibility of the candidate solution, instead of evaluating the total weight of the knapsack against the knapsack capacity, we check for the probability of satisfying the knapsack constraint. The objective of the chance constrained TTP is to find a solution (that includes a tour and a packing plan) that maximises the expected benefit while satisfying the knapsack constraint with a probability of at least $\alpha$.
Here, $\alpha$ denotes the confidence level required, which is a larger value ($\alpha\geq 0.8$). 

To estimate the probability of meeting the constraint, we refer to a weighted-scenario-based representation. Here, we consider the case where each weight $w_i$ is uniformly distributed. It should be noted that other distributions can be considered similarly, but the use of tail inequalities depends on the underlying distributions.

\section{Weighted scenario-based Model}
\label{3}
We determine the expected benefit and the feasibility of meeting the constraint by considering a limited number of different scenarios of the weight profile $\{w_1, \ldots, w_m\}$, where each scenario comes along with a certain probability of occurrence. We consider $k$ scenarios, where $T_s=\{w^s_1, \ldots, w^s_m\}$ be the $s$-th scenario, $1 \leq s \leq k$ and where the probability that a given scenario will occur $P_{T_s}$ is defined ($\sum_{s=1}^k P_{T_s} = 1.0$).
The variance of the weights in each scenario can be declared referring to certain criteria. See Table~\ref{table: weight-allocation-for-sceanrios} for an example.

For a solution $r=(x,y)$ to be feasible, we require
\begin{equation}
\label{C-for-scenario-based-model}
C(y)=\sum_{s=1}^k P_{T_s} \cdot \left( \mathbb{I}\left({\sum_{j=1}^{m}  w^s_j y_j \leq B}\right)\right) \geq \alpha.
\end{equation}

Here, the indicator $\mathbb{I}\left({\sum_{j=1}^{m}  w^s_j y_j \leq B}\right)$ returns $1$ iff $\sum_{i=1}^{m}  w^s_j y_j \leq B$ and returns $0$ otherwise. In the scenario-based model for the chance constrained TTP, we require $C(y) \geq \alpha$ for a solution to be feasible and consider $\alpha = 0.8 $ in the experiments. The objective score is calculated based on the scenarios for which the solution $y$ is feasible. For a feasible solution, we aim to maximise the expected TTP score concerning all feasible scenarios.
Hence, the goal is to maximize $E[z(x,y)] =$

\begin{equation}
\label{Z-for-scenario-based-model}
g(y) - \sum_{s=1}^k  P_{T_s} \cdot \left( \frac{d_{x_n x_1}}{\nu_{max}-\nu W^s_{x_n}} + \sum_{i=1}^{n-1} \frac{d_{x_i x_{i+1}}}{\nu_{max}-\nu W^s_{x_i}} \right),  
\end{equation}

where $W^s_{x_i}$ is the cumulative weight of the items collected from the start of the tour up to city $x_i$ in the $s$-th feasible scenario. $g(y)$ is deterministic as we assume deterministic profits.

\begin{algorithm}[t]
\caption{$\textsc{(1+1) EA}_{ws}$}
\begin{algorithmic}[1]
\STATE Input parameters: tour $x^*$, $k$ scenarios $\{T_1,~\ldots,T_k\}$.
\STATE Set the current packing plan $y = \emptyset$.
\STATE Set the best packing plan $y^* = \emptyset$.
\STATE Set best objective value $Z^*$ = $-\infty$.
\STATE Set the rate of having feasible scenarios $C=0$.
\STATE Set counter $t$= 1.
\WHILE{stopping criterion is not satisfied}
    \STATE Set $\hat{y}$ by flipping each bit in $y$ with probability $\frac{1}{n}$.
    \STATE Compute the expected objective value $Z =E[z(x^*,\hat{y})]$.
    \STATE Update $C$.
    \IF{($C \ge \alpha$ and $Z \ge Z^*$)}
        \STATE Update $y^*=\hat{y}$.
        \STATE Set $Z^*=Z$.
    \ENDIF   
    \STATE Set $t = t + 1$.
\ENDWHILE
\RETURN $y^*$
\end{algorithmic}
\label{alg: (1+1)EA_WS}
\end{algorithm}

\begin{algorithm}[t]
\caption{$\textsc{Pack}_{ws}$}
\begin{algorithmic}[1]
\STATE Input parameters: tour $x^*$, $exp$, $k$ scenarios $\{T_1,~\ldots,T_k\}$.
\STATE Compute score for each of the items $I_m \in M$.
\STATE Sort the items in non-decreasing order of scores.
\STATE Set the frequency $\omega = [m/\tau]$.
\STATE Set the current packing plan $y = \emptyset$.
\STATE Set the best packing plan $y^* = \emptyset$.
\STATE Set the best objective value $Z^* = -\infty$.
\STATE Set the rate of having feasible scenarios $C=0$.
\STATE Set the counter $t=1$ and $t^*=1$.
\WHILE{$(C>\alpha)$ and $(\omega>1)$}
\STATE Add item $I_t \in M$ to the packing plan $y=y\cup{I_t}$.
\STATE Compute $C'$.
\IF{$(C' \geq \alpha)$}
\STATE Set $C=C'$.
\IF{$(t \mod \omega =0)$}
\STATE Compute the objective value $Z=E[z(x^*,y)]$.
\IF{$(Z<Z^*)$}
\STATE Restore the packing plan $y=y^*$.
\STATE Set $t=t^*$ and update $C$.
\STATE Set $\omega=[\omega/2]$.
\ELSE
\STATE Update the best packing plan $y^*=y$.
\STATE Set $t^*=t$ and $Z^*=Z$.
\ENDIF
\ENDIF
\ELSE
\STATE Restore the packing plan $y=y^*$.
\ENDIF
\STATE Set $t=t+1$.
\ENDWHILE
\RETURN $y^*$
\end{algorithmic}

\label{alg: Pack_WS}
\end{algorithm}

\section{Algorithms}
\label{4}
After formulating the weighted scenario-based model that represents the chance constrained TTP, we work on designing algorithmic approaches to test the model. Instead of building a strategy from scratch, we select several well-known algorithms and adapt them to suit a scenario-based environment. Here, we refer to evolutionary algorithms and some heuristic approaches that have been developed to solve the classical TTP.

\subsection{Evolutionary Algorithms}
\label{4_1}
\textsc{(1+1) EA} is a structurally simple algorithm that uses a bit-wise mutation that flips each bit with a probability of $1/n$, where $n$ is the size of the solution~\cite{Droste2022}. If the mutated solution gives a better fitness score, that solution goes forward. This process continues until the stopping criteria are satisfied. To be able to use \textsc{(1+1) EA} in the scenario-based context, we mainly apply changes to the fitness evaluation and the constraints.

The modified \textsc{(1+1) EA}, namely $\textsc{(1+1) EA}_{ws}$ works as follows.
First, the algorithm takes in a TSP tour $x^*$ generated using Chained Lin-Kernighan heuristics (\textsc{CLK})~\cite{applegate2003chained} and the $k$ scenarios as input parameters. Then it initialises an empty packing plan $y$ and sets the best objective value to $-\infty$.
To keep track of the rate of having feasible scenarios, the algorithm initialises $C$ and sets it to zero. After that, the algorithm mutates the packing plan $y$ with the standard rate of $1/n$ probability and produces a new packing plan $\hat{y}$.
Then it evaluates the solution ($x^*$, $\hat{y}$) following the modified objective function (\ref{Z-for-scenario-based-model}) and obtains the expected objective value $Z$.
Along with that, the algorithm records the feasibility of the solution with each scenario and calculates $C$.
Finally, $C$ is checked against $\alpha$ to determine the feasibility of the solution. If $Z$ is better than the current best objective value $Z^*$, the algorithm updates $Z^*$ with $Z$ and takes $\hat{y}$ as the best packing plan $y^*$.
This process continues until a given stopping condition is satisfied and produces the final packing plan $y^*$, which, together with the tour $x^*$, gives the final TTP solution. 

\begin{table*}[t]
    \caption{Item weights ($w$) and probability of occurrence ($p$) of weighted-scenarios}
    \label{table: weight-allocation-for-sceanrios}
    \centering
\begin{tabular}{|c|c|c|c|c|c|}
\hline
\textbf{Scenario} & \textbf{Scenario Set A} & \textbf{Scenario Set B} & \textbf{Scenario Set C} & \textbf{Item Weights} \\
\hline
$T_1$ & $P_{T_1}=0.2$ & $P_{T_1}=0.1$ & $P_{T_1}=0.3$ & 
$ w_i =
\begin{cases} 
    a_i - \delta & \text{if } a_i \geq \delta \\
    a_i & \text{if } a_i < \delta
\end{cases}
$  \\
$T_2$ & $P_{T_2}=0.2$ & $P_{T_2}=0.1$ & $P_{T_2}=0.3$ & 
$ w_i =
\begin{cases} 
    a_i - \delta/2 & \text{if } a_i \geq \delta/2 \\
    a_i & \text{if } a_i < \delta/2
\end{cases}
$  \\
$T_3$ & $P_{T_3}=0.2$ & $P_{T_3}=0.2$ & $P_{T_3}=0.2$ & $w_i=a_i$ \\
$T_4$ & $P_{T_4}=0.2$ & $P_{T_4}=0.3$ & $P_{T_4}=0.1$ & $w_i=a_i+\delta/2$ \\
$T_5$ & $P_{T_5}=0.2$ & $P_{T_5}=0.3$ & $P_{T_5}=0.1$ & $w_i=a_i+\delta$ \\
\hline
\end{tabular}

\end{table*}

\subsection{Local Search Heuristics}
\label{4_2}
\textsc{Pack} algorithm~\cite{faulkner2015approximate} is a greedy heuristic approach that has been introduced specifically for the packing routine construction in the TTP. The \textsc{Pack} is usually run in turns for different exponent values using a re-starter algorithm called \textsc{PackIterative}~\cite{faulkner2015approximate}.
\textsc{PackIterative} has been coupled with other algorithms such as \textsc{CLK} for tour generation and has introduced an effective set of algorithms to optimise the deterministic TTP. Out of that, we chose the two highest-performing algorithms known as \textsc{S5} and \textsc{C5}.

In the \textsc{S5} arrangement, it runs the \textsc{CLK} first to generate a tour and then uses the \textsc{PackIterative} to optimise the packing plan until the solution shows no improvement. It repeats this sequence until a given stopping criterion (max time) is satisfied. In contrast, \textsc{C5} includes two more steps in this sequence called \textsc{BitFlip}~\cite{faulkner2015approximate}, and \textsc{Insertion}~\cite{faulkner2015approximate}.
The \textsc{BitFlip} attempts to further improve the packing plan by flipping each bit, while the \textsc{Inversion} tries to reduce the travel cost by delaying picking up a valuable item by modifying the tour. In summary, \textsc{C5} runs \textsc{CLK}, \textsc{PackIterative}, a single \textsc{BitFlip} step and finally an \textsc{Insertion} step in the sequence. \textsc{C5} repeats this process until it reaches the maximum runtime.

In terms of adopting these heuristic approaches to fit into the scenario-based problem definition, we mainly focus on the key component of the packing routine construction: the \textsc{Pack} algorithm, and introduce an adopted version, namely $\textsc{Pack}_{ws}$. This modification is influenced by another adopted version of the \textsc{Pack} named $\textsc{Pack}_{s}$ that has been proposed for a sampling-based representation of the chance-constrained TTP~\cite{PathirageDon2024}. Compared to the standard algorithm, modifications have been applied to the calculation of the objective value. Further, the deterministic constraint has been replaced by the chance constraint. The functionality of the adopted algorithm is as follows.

The algorithm starts by receiving a TSP tour $x^*$, $k$ scenarios and an exponent ($exp$) that will be used to compute the goodness value of items~\cite{faulkner2015approximate}.
Further, it constructs a temporary weight profile for items by averaging the weights across all scenarios. The algorithm refers to this averaged weight profile to compute the goodness value (score) for each item $I_m$. Based on the scores assigned, the algorithm sorts the items in non-decreasing order.
Then the algorithm initiates control parameters alongside an empty packing plan $y$ and sets the best objective score $Z^*$ to $-\infty$. To keep track of the rate of having feasible scenarios, the algorithm initialises $C$ and sets it to zero.
To reduce the time taken to evaluate a solution, the algorithm limits the objective score recalculation by evaluating a solution only when $\omega$ number of items have been included in a packing plan. The value of $\omega$ is calculated based on the number of items $m$ and a given integer constant $\tau$.
After that, the algorithm considers each item from the sorted list, starting with the first item $I_1 \in M$.
If the chance constraint is satisfied, the item is added to the current packing plan $y$, and the algorithm calculates the new rate of having feasible scenarios $C'$ for the updated packing plan. Then $C'$ is checked against $\alpha$ to determine the feasibility of the solution, and if satisfied, the algorithm updates $C$ with $C'$.
After adding $\omega$ number of items to $y$, the algorithm calculates the objective value $Z$. If $Z$ is less than the current best objective score $Z^*$, the value in the current packing plan $y$ is replaced by the best packing plan $y^*$.
Further, the $\omega$ is reduced to $\omega/2$, and the counter $t$ is reverted to the previous value and recalculated the $C$.
If $Z$ is better than the current best objective $Z^*$, $Z$ becomes the new best objective score, and $y$ becomes the new best packing plan $y^*$.
When no further improvement is possible or the $\omega$ reduces to 1, the algorithm stops and returns the best packing plan $y^*$. The TSP tour $x^*$ and the final packing plan $y^*$ combine to produce the final TTP solution.

\section{Experimental Investigations}
\label{5}
To investigate the performance of different algorithms, we generate multiple scenario sets referring to several benchmark instances, set experiments and analyse the results obtained.
\begin{table*}[t]
    \caption{Performance of algorithms on the weighted scenario-based model ($\alpha=0.8$)}
    \label{table: fixed-alpha_0_8}
    \resizebox{\textwidth}{!}{%
    \centering
    %\begin{tiny}
    \begin{tabular}{|c|c|c c c|c c c|c c c|}
    \hline
    \textbf{Instances} & \textbf{Algorithm} & \multicolumn{3}{|c|}{\textbf{Scenario Set A}} & \multicolumn{3}{|c|}{\textbf{Scenario Set B}} & \multicolumn{3}{|c|}{\textbf{Scenario Set C}}\\
    \cline{3-11}
    ~ & ~ & \textbf{mean} & \textbf{std} & \textbf{stat} & \textbf{mean} & \textbf{std} & \textbf{stat} & \textbf{mean} & \textbf{std} & \textbf{stat}\\
    \hline
    bsc 51 & $\textsc{(1+1) EA}_{ws}\ (1)$ & 3642.27 & 38.11 &  2(-) 3(-) & 3601.67 & 42.75 &  2(-) 3(-) & 3695.67 & 26.62 &  2(-) 3(-) \\ 
        ~ & $\textsc{S5}_{ws}\ (2)$ & 3673.13 & 0.00 &  1(+) 3(-) & 3613.85 & 0.00 &  1(+) 3(*) & 3715.02 & 0.00 &  1(+) 3(-) \\ 
        ~ & $\textsc{C5}_{ws}\ (3)$ & 3783.87 & 0.00 &  1(+) 2(+) & 3613.85 & 0.00 &  1(+) 2(*) & 3817.95 & 0.00 &  1(+) 2(+) \\ \hline
        bsc 152 & $\textsc{(1+1) EA}_{ws}\ (1)$ & 9955.52 & 211.03 &  2(-) 3(-) & 9806.39 & 204.21 &  2(-) 3(-) & 10122.09 & 219.94 &  2(*) 3(*) \\ 
        ~ & $\textsc{S5}_{ws}\ (2)$ & 10126.13 & 0.00 &  1(+) 3(*) & 10044.75 & 0.00 &  1(+) 3(*) & 10186.82 & 0.00 &  1(*) 3(*) \\ 
        ~ & $\textsc{C5}_{ws}\ (3)$ & 10126.13 & 0.00 &  1(+) 2(*) & 10044.75 & 0.00 &  1(+) 2(*) & 10186.82 & 0.00 &  1(*) 2(*) \\ \hline
        bsc 280 & $\textsc{(1+1) EA}_{ws}\ (1)$ & 16476.46 & 1038.83 &  2(-) 3(-) & 16026.35 & 1013.41 &  2(-) 3(-) & 16607.14 & 1007.49 &  2(-) 3(-) \\ 
        ~ & $\textsc{S5}_{ws}\ (2)$ & 17800.75 & 1.48 &  1(+) 3(+) & 17509.85 & 1.50 &  1(+) 3(+) & 18092.07 & 0.87 &  1(+) 3(+) \\ 
        ~ & $\textsc{C5}_{ws}\ (3)$ & 17787.89 & 12.19 &  1(+) 2(-) & 17495.69 & 14.33 &  1(+) 2(-) & 18080.35 & 8.95 &  1(+) 2(-) \\ \hline
        bsc 575 & $\textsc{(1+1) EA}_{ws}\ (1)$ & 28611.86 & 432.26 &  2(-) 3(-) & 27989.57 & 521.98 &  2(-) 3(-) & 29085.65 & 290.39 &  2(-) 3(-) \\ 
        ~ & $\textsc{S5}_{ws}\ (2)$ & 30401.24 & 213.60 &  1(+) 3(*) & 29997.07 & 189.50 &  1(+) 3(*) & 30996.53 & 101.56 &  1(+) 3(+) \\ 
        ~ & $\textsc{C5}_{ws}\ (3)$ & 30047.68 & 434.58 &  1(+) 2(*) & 29865.23 & 251.03 &  1(+) 2(*) & 30776.91 & 291.58 &  1(+) 2(-) \\ \hline
        bsc 1000 & $\textsc{(1+1) EA}_{ws}\ (1)$ & 135452.80 & 242.27 &  2(+) 3(+) & 135388.90 & 308.99 &  2(+) 3(+) & 135491.00 & 179.36 &  2(+) 3(+) \\ 
        ~ & $\textsc{S5}_{ws}\ (2)$ & 129849.87 & 460.28 &  1(-) 3(*) & 130417.00 & 263.51 &  1(-) 3(+) & 130157.43 & 546.07 &  1(-) 3(+) \\ 
        ~ & $\textsc{C5}_{ws}\ (3)$ & 130095.60 & 351.99 &  1(-) 2(*) & 129797.27 & 555.96 &  1(-) 2(-) & 128904.20 & 658.51 &  1(-) 2(-) \\ \hline
        unc 51 & $\textsc{(1+1) EA}_{ws}\ (1)$ & 1955.86 & 166.39 &  2(-) 3(-) & 1846.01 & 137.57 &  2(-) 3(-) & 2000.79 & 178.18 &  2(-) 3(-) \\ 
        ~ & $\textsc{S5}_{ws}\ (2)$ & 2120.35 & 0.00 &  1(+) 3(*) & 1984.74 & 60.66 &  1(+) 3(-) & 2269.16 & 0.00 &  1(+) 3(*) \\ 
        ~ & $\textsc{C5}_{ws}\ (3)$ & 2120.35 & 0.00 &  1(+) 2(*) & 2011.24 & 0.00 &  1(+) 2(+) & 2269.16 & 0.00 &  1(+) 2(*) \\ \hline
        unc 152 & $\textsc{(1+1) EA}_{ws}\ (1)$ & 3459.73 & 1052.84 &  2(-) 3(-) & 3166.32 & 1032.54 &  2(-) 3(-) & 3949.38 & 1068.58 &  2(-) 3(-) \\ 
        ~ & $\textsc{S5}_{ws}\ (2)$ & 4727.69 & 0.20 &  1(+) 3(-) & 4410.60 & 0.19 &  1(+) 3(-) & 5098.03 & 0.16 &  1(+) 3(-) \\ 
        ~ & $\textsc{C5}_{ws}\ (3)$ & 5259.81 & 0.12 &  1(+) 2(+) & 4939.83 & 0.15 &  1(+) 2(+) & 5587.61 & 0.12 &  1(+) 2(+) \\ \hline
        unc 280 & $\textsc{(1+1) EA}_{ws}\ (1)$ & 16974.81 & 1193.24 &  2(-) 3(-) & 16062.29 & 1239.22 &  2(-) 3(-) & 17762.58 & 1249.64 &  2(-) 3(-) \\ 
        ~ & $\textsc{S5}_{ws}\ (2)$ & 18423.40 & 51.67 &  1(+) 3(-) & 17816.56 & 48.65 &  1(+) 3(-) & 19140.82 & 1.29 &  1(+) 3(-) \\ 
        ~ & $\textsc{C5}_{ws}\ (3)$ & 18820.63 & 43.48 &  1(+) 2(+) & 17828.66 & 13.13 &  1(+) 2(+) & 19543.47 & 23.92 &  1(+) 2(+) \\ \hline
        unc 575 & $\textsc{(1+1) EA}_{ws}\ (1)$ & 31260.68 & 613.44 &  2(-) 3(-) & 30008.96 & 594.06 &  2(-) 3(-) & 32852.93 & 582.52 &  2(-) 3(-) \\ 
        ~ & $\textsc{S5}_{ws}\ (2)$ & 32585.62 & 384.27 &  1(+) 3(*) & 32323.57 & 379.72 &  1(+) 3(+) & 35184.45 & 360.24 &  1(+) 3(+) \\ 
        ~ & $\textsc{C5}_{ws}\ (3)$ & 32470.94 & 249.50 &  1(+) 2(*) & 31626.19 & 405.54 &  1(+) 2(-) & 34398.46 & 437.51 &  1(+) 2(-) \\ \hline
        unc 1000 & $\textsc{(1+1) EA}_{ws}\ (1)$ & 154819.03 & 267.37 &  2(+) 3(*) & 154788.17 & 235.72 &  2(+) 3(+) & 154904.93 & 204.76 &  2(+) 3(+) \\ 
        ~ & $\textsc{S5}_{ws}\ (2)$ & 124294.13 & 67.31 &  1(-) 3(-) & 154472.50 & 156.84 &  1(-) 3(+) & 154311.53 & 193.03 &  1(-) 3(+) \\ 
        ~ & $\textsc{C5}_{ws}\ (3)$ & 154693.67 & 192.60 &  1(*) 2(+) & 154117.30 & 296.94 &  1(-) 2(-) & 153950.47 & 114.92 &  1(-) 2(-) \\ \hline
    \end{tabular}
   % \end{tiny}
}

\end{table*}
\begin{table*}[t]
    \caption{Performance of algorithms on the weighted scenario-based model ($\alpha=0.9$)}
    \label{table: fixed-alpha_0_9}
    \resizebox{\textwidth}{!}{%
    \centering
    %\begin{tiny}
    \begin{tabular}{|c|c|c c c|c c c|c c c|}
    \hline
    \textbf{Instances} & \textbf{Algorithm} & \multicolumn{3}{|c|}{\textbf{Scenario Set A}} & \multicolumn{3}{|c|}{\textbf{Scenario Set B}} & \multicolumn{3}{|c|}{\textbf{Scenario Set C}}\\
    \cline{3-11}
    ~ & ~ & \textbf{mean} & \textbf{std} & \textbf{stat} & \textbf{mean} & \textbf{std} & \textbf{stat} & \textbf{mean} & \textbf{std} & \textbf{stat}\\
    \hline
    bsc 51 & $\textsc{(1+1) EA}_{ws}\ (1)$ & 3646.71 & 36.03 &  2(-) 3(-) & 3604.29 & 40.60 &  2(-) 3(-) & 3692.53 & 27.07 &  2(-) 3(-) \\ 
        ~ & $\textsc{S5}_{ws}\ (2)$ & 3673.13 & 0.00 &  1(+) 3(-) & 3613.85 & 0.00 &  1(+) 3(*) & 3715.02 & 0.00 &  1(+) 3(-) \\ 
        ~ & $\textsc{C5}_{ws}\ (3)$ & 3783.87 & 0.00 &  1(+) 2(+) & 3613.85 & 0.00 &  1(+) 2(*) & 3817.95 & 0.00 &  1(+) 2(+) \\ \hline
        bsc 152 & $\textsc{(1+1) EA}_{ws}\ (1)$ & 10044.38 & 188.65 &  2(*) 3(*) & 9862.30 & 286.40 &  2(*) 3(*) & 10172.59 & 152.46 &  2(+) 3(*) \\ 
        ~ & $\textsc{S5}_{ws}\ (2)$ & 10126.13 & 0.00 &  1(*) 3(*) & 10044.75 & 0.00 &  1(*) 3(*) & 10008.69 & 21.88 &  1(-) 3(-) \\ 
        ~ & $\textsc{C5}_{ws}\ (3)$ & 10126.13 & 0.00 &  1(*) 2(*) & 10044.75 & 0.00 &  1(*) 2(*) & 10230.36 & 0.00 &  1(*) 2(+) \\ \hline
        bsc 280 & $\textsc{(1+1) EA}_{ws}\ (1)$ & 16322.84 & 970.75 &  2(-) 3(-) & 16128.36 & 1045.87 &  2(-) 3(-) & 16618.71 & 979.50 &  2(-) 3(-) \\ 
        ~ & $\textsc{S5}_{ws}\ (2)$ & 17800.86 & 2.06 &  1(+) 3(+) & 17509.75 & 1.51 &  1(+) 3(+) & 18091.91 & 1.20 &  1(+) 3(+) \\ 
        ~ & $\textsc{C5}_{ws}\ (3)$ & 17778.26 & 14.82 &  1(+) 2(-) & 17494.77 & 14.71 &  1(+) 2(-) & 18078.67 & 9.47 &  1(+) 2(-) \\ \hline
        bsc 575 & $\textsc{(1+1) EA}_{ws}\ (1)$ & 28650.36 & 596.23 &  2(-) 3(-) & 27785.31 & 519.53 &  2(-) 3(-) & 29258.23 & 553.13 &  2(-) 3(-) \\ 
        ~ & $\textsc{S5}_{ws}\ (2)$ & 30481.73 & 155.04 &  1(+) 3(*) & 29946.83 & 124.08 &  1(+) 3(*) & 30952.07 & 231.17 &  1(+) 3(*) \\ 
        ~ & $\textsc{C5}_{ws}\ (3)$ & 30356.08 & 260.92 &  1(+) 2(*) & 29688.34 & 353.34 &  1(+) 2(*) & 30761.77 & 297.21 &  1(+) 2(*) \\ \hline
        bsc 1000 & $\textsc{(1+1) EA}_{ws}\ (1)$ & 135484.17 & 259.92 &  2(+) 3(+) & 135490.13 & 241.29 &  2(+) 3(+) & 135553.83 & 244.57 &  2(+) 3(+) \\ 
        ~ & $\textsc{S5}_{ws}\ (2)$ & 130700.00 & 181.92 &  1(-) 3(+) & 130368.13 & 232.05 &  1(-) 3(+) & 129850.70 & 301.72 &  1(-) 3(*) \\ 
        ~ & $\textsc{C5}_{ws}\ (3)$ & 129806.33 & 477.09 &  1(-) 2(-) & 129851.40 & 506.86 &  1(-) 2(-) & 130181.80 & 415.34 &  1(-) 2(*) \\ \hline
        unc 51 & $\textsc{(1+1) EA}_{ws}\ (1)$ & 1945.50 & 165.07 &  2(-) 3(-) & 1817.64 & 137.31 &  2(-) 3(-) & 2115.78 & 190.25 &  2(-) 3(-) \\ 
        ~ & $\textsc{S5}_{ws}\ (2)$ & 2120.35 & 0.00 &  1(+) 3(*) & 1973.67 & 0.00 &  1(+) 3(-) & 2269.16 & 0.00 &  1(+) 3(*) \\ 
        ~ & $\textsc{C5}_{ws}\ (3)$ & 2120.35 & 0.00 &  1(+) 2(*) & 2011.24 & 0.00 &  1(+) 2(+) & 2269.16 & 0.00 &  1(+) 2(*) \\ \hline
        unc 152 & $\textsc{(1+1) EA}_{ws}\ (1)$ & 3936.32 & 1024.53 &  2(-) 3(-) & 3209.76 & 1048.03 &  2(-) 3(-) & 4363.37 & 1026.42 &  2(-) 3(-) \\ 
        ~ & $\textsc{S5}_{ws}\ (2)$ & 4727.63 & 0.18 &  1(+) 3(-) & 4410.66 & 0.21 &  1(+) 3(-) & 5098.05 & 0.17 &  1(+) 3(-) \\ 
        ~ & $\textsc{C5}_{ws}\ (3)$ & 5259.79 & 0.13 &  1(+) 2(+) & 4939.83 & 0.15 &  1(+) 2(+) & 5587.61 & 0.12 &  1(+) 2(+) \\ \hline
        unc 280 & $\textsc{(1+1) EA}_{ws}\ (1)$ & 16663.89 & 1203.70 &  2(-) 3(-) & 16093.65 & 1188.16 &  2(-) 3(-) & 17358.51 & 1250.55 &  2(-) 3(-) \\ 
        ~ & $\textsc{S5}_{ws}\ (2)$ & 18424.15 & 51.38 &  1(+) 3(-) & 17803.96 & 1.25 &  1(+) 3(-) & 19168.25 & 1.10 &  1(+) 3(-) \\ 
        ~ & $\textsc{C5}_{ws}\ (3)$ & 18684.13 & 51.04 &  1(+) 2(+) & 17830.95 & 9.63 &  1(+) 2(+) & 19561.19 & 19.26 &  1(+) 2(+) \\ \hline
        unc 575 & $\textsc{(1+1) EA}_{ws}\ (1)$ & 31207.53 & 843.47 &  2(-) 3(-) & 30046.32 & 628.59 &  2(-) 3(-) & 32625.82 & 735.23 &  2(-) 3(-) \\ 
        ~ & $\textsc{S5}_{ws}\ (2)$ & 33640.65 & 393.70 &  1(+) 3(+) & 32259.60 & 348.07 &  1(+) 3(+) & 34047.98 & 144.32 &  1(+) 3(+) \\ 
        ~ & $\textsc{C5}_{ws}\ (3)$ & 33002.21 & 508.82 &  1(+) 2(-) & 31634.72 & 300.10 &  1(+) 2(-) & 33789.53 & 341.30 &  1(+) 2(-) \\ \hline
        unc 1000 & $\textsc{(1+1) EA}_{ws}\ (1)$ & 154767.40 & 279.36 &  2(+) 3(+) & 154819.10 & 295.90 &  2(*) 3(+) & 154881.30 & 246.07 &  2(+) 3(*) \\ 
        ~ & $\textsc{S5}_{ws}\ (2)$ & 154446.53 & 176.03 &  1(-) 3(+) & 154577.87 & 157.42 &  1(*) 3(+) & 139709.19 & 176.18 &  1(-) 3(-) \\ 
        ~ & $\textsc{C5}_{ws}\ (3)$ & 153954.03 & 297.83 &  1(-) 2(-) & 154093.30 & 276.02 &  1(-) 2(-) & 154715.07 & 103.61 &  1(*) 2(+) \\ \hline
    \end{tabular}
   % \end{tiny}
}

\end{table*}

\subsection{Experimental Setup}
\label{5_1}

We use TTP benchmark instances from ~\cite{polyakovskiy2014comprehensive} in our experiments.
The size of the instances that we use ranges from $51$ to $1000$ cities with a single item allocated to each city, except the first city. The total of $10$ instances that are selected consists of $5$ instances with \textit{uncorrelated} weights (unc) and $5$ instances with \textit{bounded strongly correlated} weights (bsc).

To create scenarios, we refer to the weight profile as per the standard TTP instances and generate new weight profiles. The weights of the items in scenarios are based on a predefined setup that slightly adjusts the original weights (Table~\ref{table: weight-allocation-for-sceanrios}). 
For that, we use a small value ($\delta$) to change the weights ($a_i$) in the standard TTP instance. The changes are applied in a certain pattern so that the weights from Scenario $1$ to Scenario $5$ mostly stay in order. 
In Scenario $1$, the standard weight is reduced by $\delta$ amount and in Scenario $2$, the weight is reduced by $\delta/2$. In both of these scenarios, we keep the standard weight as it is if $\delta$ is greater than the weight $a_i$, to make sure that the weight profiles do not contain any negative values. Scenario $3$ uses the standard weight as it is. Scenario $4$ uses the weight increased by $\delta/2$ amount. Finally, Scenario $5$ uses the standard weight increased by $\delta$.  In our experimental setting, we use $\delta=20$.
Referring to this weight setup, we define three different sets of scenarios (Scenario Set A, B and C), with changes applied to the probability of occurrence ($P_T$) in each scenario as in Table~\ref{table: weight-allocation-for-sceanrios}. The Scenario Set A represents where all individual scenarios share a similar probability of occurrence. The Scenario Set B represents a case where the scenarios with higher weights have a high probability of occurrence. In contrast, the Scenario Set C represents a case where the scenarios with lower weights have a high probability of occurrence. Each set includes $5$ distinct scenarios ($k=5$).

We run each algorithm for a maximum of 10 minutes. Further, we run each experiment independently 30 times to obtain the mean objective score and the standard deviation for the analysis. We test the entire experimental setup for two different $\alpha$ values ($\alpha \in \{0.8, 0.9\}$), where $\alpha$ is the threshold value for the chance constraint.

Further, we conduct statistical tests to confirm the statistical significance of the distribution of the objective scores. In the beginning, we use the Kruskal-Wallis test to determine if at least one of the algorithms has a significantly different distribution. If the test notes significance, then we conduct the Dunn test using Bonferroni correction for the p-values to do a pair-wise comparison among the three algorithms.

\subsection{Experimental Results}
\label{5_2}

We represent the results along with the statistical tests in two tables (Table~\ref{table: fixed-alpha_0_8} and Table~\ref{table: fixed-alpha_0_9}), which mainly bring a comparison between the performance of different algorithms in each scenario set.
The column \textit{mean} in the tables represents the mean objective score obtained for 30 independent runs, and the column \textit{std} denotes its standard deviation. The column \textit{stat} represents the statistical test results that compare the three algorithms relating to each scenario set in each instance. For example, under the Scenario Set  A, $2(-)\ 3(-)$ given in the \textit{stat} column for bsc $51$ instance in the first row of the Table~\ref{table: fixed-alpha_0_8} denotes, the current algorithm $\textsc{(1+1) EA}_{ws}\ (1)$ is statistically significantly worse than $\textsc{S5}_{ws}\ (2)$ and $\textsc{C5}_{ws}\ (3)$. Similarly, the $(+)$ sign used in the table denotes that the current algorithm is statistically significantly better than the algorithm it is compared with, and $(*)$ states there is no significant difference.

Table~\ref{table: fixed-alpha_0_8} represents the results obtained for $\alpha=0.8$.
For the smaller instances, the two heuristic algorithms perform better than the evolutionary algorithm, and that is observable throughout all scenario sets. In comparison, for larger instances (bsc $1000$ and unc $1000$), the performance of $\textsc{(1+1) EA}_{ws}\ (1)$ brings slightly better performance than the heuristic methods. A possible reason for this observation would be the higher number of iterations the $\textsc{(1+1) EA}_{ws}\ (1)$ can handle within the given time bound, compared to the heuristic algorithms. Since it takes a comparatively longer time to see a convergence in the larger instances, the higher number of iterations may provide an advantage. Another interesting observation in the larger instances is that the $\textsc{S5}_{ws}$'s performance in Scenario Set A is significantly lower, yet the $\textsc{C5}_{ws}$ manages to gain a comparatively better performance. This means that even if the focused optimisation of the packing plan is effective in most situations, when the size of the tour is larger, it is beneficial to influence the tour component to gain better optimisation. This is only visible in Scenario Set A as the scenarios are assigned with a similar probability of occurrence ($P_T$). The effect of the $P_T$ is varied in Scenario Set B and Scenario Set C.

Scenario Set B represents a situation where there is a higher probability of having larger weights for the items. That would essentially reduce the expected objective value, as there is a higher chance of violating the chance constraint. In contrast, Scenario Set C represents a situation where there is a higher probability of having smaller weights for the items. Here, it should gain the expected objective value as it leaves the agent (the thief) to gain more profit by collecting comparatively lesser weights. This relationship between the two scenarios is depicted in the results as Scenario Set B produces comparatively the lowest expected objective values, while Scenario Set C gives the best scores.

Table~\ref{table: fixed-alpha_0_9} represents the results obtained for the threshold value $\alpha$ set to $0.9$. When the threshold value is getting larger, we expect the objective scores to become lower as it tightens the chance constraint further. That is articulated through the overall results, except for a few exceptions. Comparatively, the results obtained for $\alpha=0.9$ show higher standard deviations, which can be a possible result of having a tighter chance constraint. Again, the results for larger instances can be affected by the algorithmic structures as discussed above. Overall, the algorithm $\textsc{S5}_{ws}$ performs better than the evolutionary algorithm. $\textsc{C5}_{ws}$ shows equal or narrowly better improvements compared to $\textsc{S5}_{ws}$. However, $\textsc{S5}_{ws}$ still demonstrate effectiveness when comparing the complexity of the two algorithms against the results they produce. The statistical significance among algorithms confirms these observations.

\section{Conclusions}
In this paper, we represented the chance constrained TTP using a weighted scenario-based model and produced three adopted versions of popular algorithms to optimise the stochastic TTP instances in a scenario-based representation.
Overall, the heuristic algorithms show better performance and convergence compared to the evolutionary approach.
$\textsc{S5}_{ws}$ can be recognised as an effective algorithm among the algorithms we adopted.
The outcome shows the effectiveness of employing a weighted scenario-based model as a relaxed strategy to handle stochastic multi-component problems that involve chance constraints.

\section*{Acknowledgment}
\label{6}
This work was supported by the Australian Research Council (ARC) through grant FT200100536 and with super-computing resources provided by the Phoenix HPC service at the University of Adelaide.

\bibliographystyle{unsrt}
\bibliography{main}
\end{document}